\documentclass[runningheads]{llncs}

\usepackage{eccv}

\usepackage{eccvabbrv}

\usepackage{graphicx}
\usepackage{booktabs}
\usepackage{multirow}
\usepackage{wrapfig}

\usepackage[accsupp]{axessibility}  

\usepackage{algorithm}
\usepackage{algpseudocode}
\usepackage{xcolor}
\usepackage[table]{xcolor}
\algrenewcommand\algorithmiccomment[1]{\hfill{\color{gray}\footnotesize #1}}

\usepackage[pagebackref,breaklinks,colorlinks,citecolor=eccvblue]{hyperref}
\usepackage{hyperref}

\usepackage{orcidlink}

\def\modelname{UniFS}

\begin{document}

\title{\modelname: Unified Fast-to-Slow Hierarchical Architecture for Vision-Language-Action Models}
\titlerunning{\modelname}

\author{Lin Sun\inst{1,2} \and
Zhiwei Guan\inst{1} \and Conglin Wang\inst{1} \and Zihong Chen\inst{1} \and Jianhai Yu\inst{1} \and Zongsheng Li\inst{1} \and Boyong He\inst{1} \and Tao Sun\inst{3} \and Jiale Cao\inst{2}$^{*}$ \and Lige Liu\inst{1}$^{*}$}

\authorrunning{L Sun~et al.}
\institute{JD.com, China \and
Tianjin University, China \and 
Yiwu Research Institute of Fudan University, China\\
\email{\{sunlin.112,~wangconglin.11,~liulige.neo\}@jd.com} \\
\email{connor@tju.edu.cn}}

\footnotetext{* Corresponding authors.}

\maketitle
\begin{abstract}
Mainstream Fast-Slow dual system vision-language-action models decouple a high-frequency action expert from a low-frequency vision-language model for efficiency, yet they face a fundamental frequency dilemma: large update gaps cause semantic drift from stale context, while small gaps erode the intended computational savings. Moreover, because the action expert receives only the VLM's final-layer representation at a single fixed frequency, rich intermediate features are discarded, limiting both information coupling and manipulation precision. Inspired by multi-timescale neural processing in the human brain, we introduce \modelname, a unified fast-to-slow architecture that resolves these challenges through three key designs. First, we stratify the VLM layers into groups with progressively decreasing update frequencies, enabling shallow layers to capture fast-changing dynamics while deeper layers cache stable semantic context. Second, a latent vector inversion mechanism re-routes the interaction order between multi-scale VLM features and the action expert, aligning fast-varying representations with fine-grained action decoding and slow-varying ones with coarse planning. Third, a multi-level supervision strategy enforces a coarse-to-fine learning hierarchy across temporal scales. Together, these designs enable richer cross-frequency information transfer within a single backbone, while the low-frequency pathways additionally preserve temporal context across steps. Experiments on LIBERO show that \modelname\ achieves state-of-the-art performance ($98.3\%$ average success rate, a $2.5\%$ gain over VLA-Adapter baseline) while reducing average inference latency from $36.5$ ms to $17.8$ ms ($2.1$x speedup). Real-robot experiments on a Franka platform further validate its practical applicability. Code is opensourced at \url{https://github.com/linsun449/UniFS}.

  \keywords{VLA \and Fast-to-Slow System \and Multi-timescale Representation}
\end{abstract}    
\section{Introduction}
\label{sec:intro}

Vision-Language-Action (VLA) models~\cite{jiang2025better, tan2025think, black2024pi0, zheng2025x, zhao2025cot, shi2025memoryvla, li2025sp, li2025cogvla, zitkovich2023rt, kim2024openvla, yue2024deer, bjorck2025gr00t} have emerged as a cornerstone for generalist Embodied AI, offering remarkable zero-shot generalization for complex manipulation tasks. Recent works, including $\pi_0$~\cite{black2024pi0}, $\pi_{0.5}$~\cite{intelligence2025pi_}, Hume~\cite{song2025hume}, CogVLA~\cite{li2025cogvla}, Deer-VLA~\cite{yue2024deer}, and GraspVLA~\cite{deng2025graspvla}, further advance this paradigm by moving to continuous action generation for fine-grained control. However, they inherit a critical bottleneck from their large-scale Vision-Language Model (VLM) backbones, which already introduce substantial latency for a single forward pass~\cite{ma2024survey, wen2025tinyvla, yue2024deer}, resulting in end-to-end control frequencies of only 1–10 Hz, far below the levels typically required for stable closed-loop real-time control in dynamic environments~\cite{hu2023toward, chignoli2021humanoid, chen2024image}.

\begin{figure*}[t]
    \centering
    \includegraphics[width=1\textwidth]{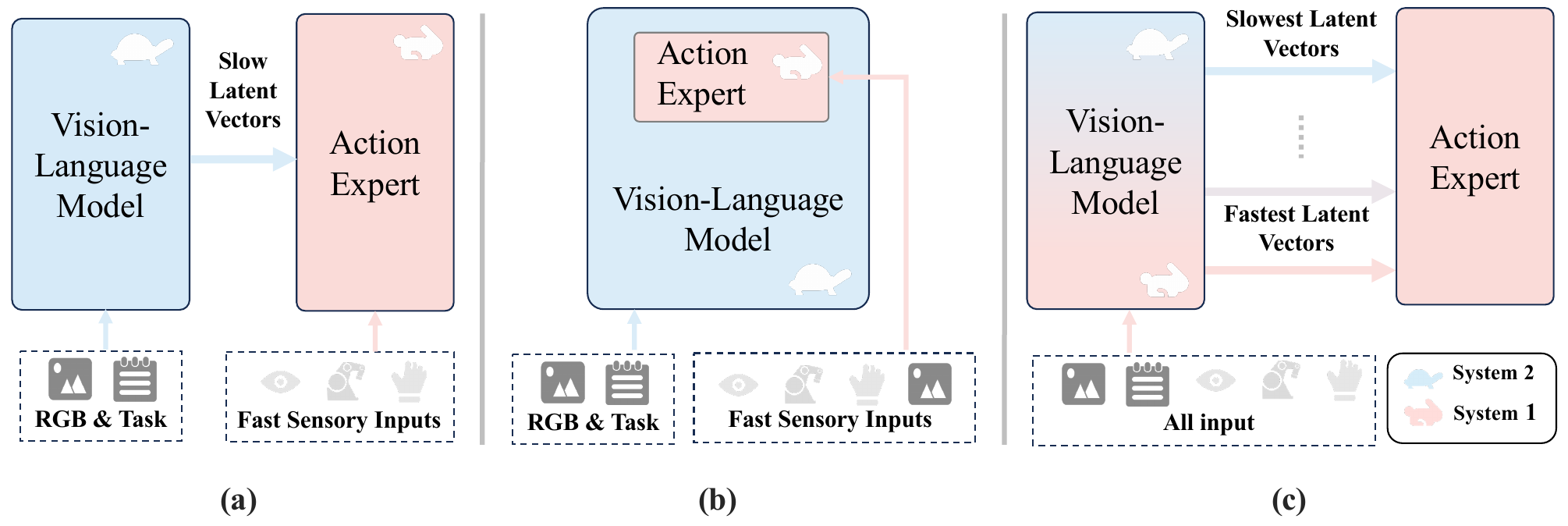}
    \caption{Motivation and Architecture Overview. (a) The separate-module paradigm transmits only slow latent vectors from the VLM to action expert. (b) The embedded paradigm integrates the action expert directly 
    within the VLM layers. (c) Mainstream methods suffer from rigid hard-linking between separate modules. Our approach unifies these dynamics via multi-frequency latent vector for seamless coordination.}
    \label{fig:01intro}
    \vspace{-15pt}
\end{figure*}

To mitigate this latency, recent works have explored dual-system architectures~\cite{bjorck2025gr00t, cui2025openhelix, chen2025fast, han2024dual, song2025hume, li2025sp}, decoupling high-level planning from low-level control at different frequencies. As illustrated in Fig.~\ref{fig:01intro}(a), one line of works, represented by GR00T-N1~\cite{bjorck2025gr00t}, maintains the VLM and action expert as separate modules, where the VLM periodically transmits slow latent representations to guide the fast action expert. Another line of works, represented by 
Fast-in-Slow~\cite{chen2025fast}, embeds the action expert directly within the VLM, as shown in Fig.~\ref{fig:01intro}(b), enabling tighter coupling between the two systems. Despite their architectural differences, both paradigms impose significant constraints on the VLM-action expert interaction: the action expert relies on the final-layer latent representation, discarding rich multi-scale semantic features, while the fixed exchange frequency further undermines adaptive responses to dynamic task demands~\cite{cui2025openhelix}.

\begin{figure*}[t]
    \centering
    \includegraphics[width=1\textwidth]{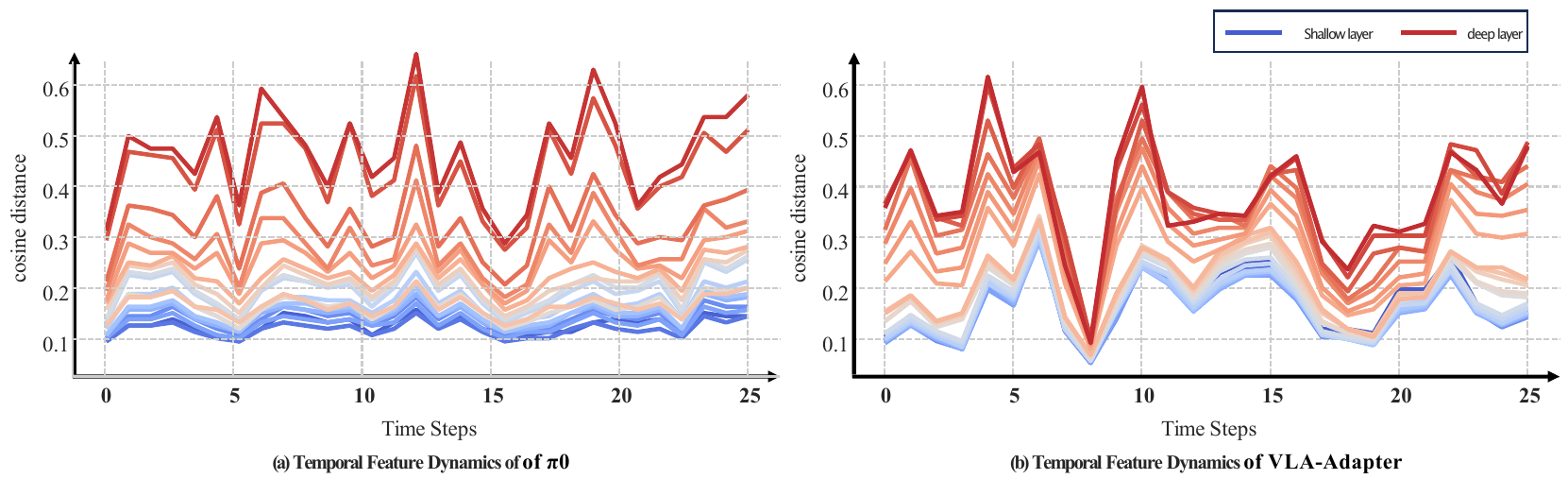}
    \caption{\textbf{Cosine distance} between layer-wise latent representations over time. (a) Layer-wise feature evolution in the $\pi_0$ model over time steps. (b) Corresponding temporal feature dynamics in the VLA-Adapter. The curves represent different network layers, illustrating how feature stability varies across the depth of the models.}
    \label{fig:insight}
    \vspace{-10pt}
\end{figure*}

To address these challenges, we draw inspiration from the multi-timescale processing of the human brain \cite{kahneman2011thinking, behrouz2025nested}, where communication between the cerebral cortex and cerebellum leverages multiple frequency bands (e.g., 30–150 Hz gamma waves for sensory information and 13–30 Hz beta waves for higher-level cognitive processing) to transmit and integrate information simultaneously. Motivated by this principle, we visualize how the latent representations at each layer in $\pi_0$ and the VLA-Adapter change over time, as shown in \Cref{fig:insight}, and find that the magnitude of these changes differs substantially across layers, a phenomenon that aligns with the brain’s multiple temporal scales.

These observations and neurobiological perspective motivates a design principle that instead of tying frequency to specific modules, we vary it smoothly across the whole model. In this way, we introduce \modelname, as shown in \Cref{fig:01intro}(c), a novel unified fast-to-slow architecture that directly generates multi-timescale latent representations from the VLM. These multi-timescale latent representations further support the action expert in generating actions. 

This unified design integrates fast and slow dynamics within a single VLM, leading to more efficient parameter utilization. By feeding multi-frequency latent features produced by the VLM into the action expert, the proposed design substantially improves the alignment of representation spaces between the VLM and the action expert, thereby effectively alleviating the insufficient information coupling between them. Moreover, by establishing multi-frequency pathways from the VLM to the action expert, the proposed architecture helps mitigate the frequency dilemma that commonly arises in mainstream fast–slow dual systems. During inference, slow-frequency latent representations do not need to be updated at every step and can be cached, enabling substantial acceleration of the overall inference process without sacrificing performance. In addition, the slow-frequency pathways endow the model with an implicit memory mechanism that intrinsically preserves historical dependencies without requiring explicit memory buffers, a capability that is fundamentally absent in mainstream fast–slow systems.

Our contributions are summarized as follows:
\begin{itemize}
    \item We propose a novel unified fast-to-slow hierarchical architecture that directly provides multi-timescale latent representations to the action expert from a single VLM, thereby addressing the frequency dilemma in mainstream fast–slow dual systems and improving the alignment between visual, language and action representations.
    \item We demonstrate that this architecture achieves substantial inference-time acceleration without sacrificing performance, while simultaneously endowing the model with implicit memory capabilities. In particular, \modelname\ achieves a 2.5\% performance improvement on LIBERO while reducing the average inference latency from 36.5\,ms to 17.8\,ms.
    \item We conduct extensive experiments to validate the effectiveness of \modelname\ in terms of efficiency and robustness. The code will be released to facilitate further research.

\end{itemize}

\section{Related Work}

\subsection{The Rise of Large-Scale VLAs and Efficiency Challenges} 
Recent advances in Embodied AI have been driven by VLA models, which leverage internet-scale pretraining to achieve unprecedented zero-shot generalization. Seminal works like RT-2~\cite{zitkovich2023rt} and OpenVLA~\cite{kim2024openvla} have demonstrated the power of unifying natural language reasoning with robotic control, while newer systems (e.g.  $\pi_{0.5}$~\cite{intelligence2025pi_}, X-VLA~\cite{zheng2025x}, MemoryVLA~\cite{shi2025memoryvla}) have further enhanced precision by shifting from discrete tokens to continuous action spaces via iterative decoding (e.g. diffusion~\cite{wen2024diffusion} or flow matching~\cite{intelligence2025pi_, bjorck2025gr00t} ). However, this surge in capability introduces a critical efficiency bottleneck. The massive computational overhead of large VLM backbones (typically 7B+ parameters), compounded by the multi-step sampling required for continuous action generation, restricts inference frequencies to merely 5–10 Hz. This falls drastically short of the >50 Hz threshold required for stable real-time control in dynamic environments. Consequently, the architecture of these VLAs has become a fundamental barrier to their practical deployment, necessitating novel architectural paradigms that can decouple computational complexity from control frequency.

\subsection{Fast-Slow Dual-Systems} 

To address the efficiency bottleneck imposed by large-parameter VLA models, the Fast-Slow Dual-System paradigm has emerged as the mainstream solution for real-time control. Inspired by Kahneman's Dual-Process Theory~\cite{kahneman2011thinking}, this approach leverages frequency differentiation to balance semantic reasoning with real-time execution. Representative works, such as HiRT~\cite{zhang2024hirt}, RoboDual~\cite{bu2024towards}, and OpenHelix~\cite{cui2025openhelix}, employ asynchronous strategies where a large VLM serves as a low-frequency Slow System (1–5 Hz) for high-level planning, while a lightweight policy acts as a high-frequency Fast System (20–50 Hz) for reactive motor control.
While effective in reducing average computational load, these explicit dual-module implementations suffer from inherent structural limitations.
Specifically, reliance on rigid hard-coupling strategies with fixed synchronization rates creates a critical trade-off: large frequency gaps induce semantic drift due to outdated context, whereas small gaps negate acceleration benefits. Furthermore, the disjoint nature of these architectures hinders true end-to-end optimization, as gradients cannot flow seamlessly between independent planning and control modules, often leading to suboptimal feature alignment. Additionally, existing methods typically employ shallow fusion mechanisms (e.g., simple concatenation) at module boundaries, failing to capture rich multi-scale temporal dependencies within the latent space. 

\subsection{Towards Bio-Inspired Multi-Frequency Latent Fusion}
These structural flaws point towards a need for unified architectures. Neuroscience offers a compelling blueprint: the brain processes multi-timescale information not via segregated modules, but through multi-frequency neural oscillations coexisting within unified circuits~\cite{snyder2026resonant, senkowski2024multi, jerath2019hierarchical, shine2019neuromodulatory}. While some AI works have explored bio-inspired designs~\cite{sapkota2025vision, liu2025neural}, few have translated this into frequency-aware latent fusion for end-to-end VLAs~\cite{yu2025survey, zhang2025pure}. Most existing strategies treat latents as static features, ignoring their spectral properties. In contrast, our work proposes \modelname\, which embeds frequency-specific pathways directly into a single Transformer backbone. This enables deep fusion of fast dynamics and slow semantics within the latent space, overcoming the fragmentation of explicit dual-systems while preserving the generalization of large foundation models.
\section{Method}
\label{sec:method}
\subsection{Problem Formulation}
\label{subsec:problem}
VLA models enable robots to generate action sequences by reasoning jointly over environmental observations and natural language instructions. Specifically, given a language instruction $\mathbf{L}$ and a set of visual observations $\mathbf{O}_{\text{vis}} \in \mathbb{R}^{H \times W \times 3}$, the VLA policy $\pi_\theta$ generates an action chunk $\mathbf{A} \in \mathbb{R}^{T \times D}$ for robotic manipulation, denoted as:
\begin{equation}
\mathbf{A} = [\mathbf{a}_1, \mathbf{a}_2, \ldots, \mathbf{a}_T] = \pi_\theta(\mathbf{O}_{\text{vis}}, \mathbf{L}_{\text{task}})
\end{equation}
where $T$ denotes the chunk length, $D$ represents the action dimension of the specific robot, and each action step $\mathbf{a}_i \in \mathbb{R}^{D}$ serves as a control command. In our implementation, we utilize a 7-DoF action representation:
\begin{equation}
\mathbf{a}_i = [\Delta x, \Delta y, \Delta z, \Delta \phi, \Delta \theta, \Delta \psi, g]^{\top}
\end{equation}
where $\Delta x, \Delta y, \Delta z$ denote the relative end-effector translation offsets, $\Delta \phi, \Delta \theta, \Delta \psi$ denote the rotation changes (roll, pitch, yaw), and $g \in \{0, 1\}$ indicates the gripper state (open/closed). Accordingly, the action dimension is $D=7$.

The conventional fast-slow dual system implements the policy $\pi_\theta$ by decomposing the process into two distinct frequency. At the slow-frequency scale $f_{\text{slow}}$, visual observations and language instructions are processed by the VLM to produce a slow latent representation $\mathbf{v}_{\text{slow}}$, formulated as:
\begin{equation}
    \mathbf{v}_{\text{slow}} = \pi_{\text{VLM}}(\mathbf{O}_{\text{vis}}, \mathbf{L}).
\end{equation}

At the fast-frequency scale $f_{\text{fast}}$, this latent representation $\mathbf{v}_{\text{slow}}$ is held fixed for multiple steps and combined with fast sensory inputs $\mathbf{O}_{\text{sen}}$ and proprioceptive states $\mathbf{O}_{\text{prop}}$ for action execution, denoted by:
\begin{equation}
    \mathbf{A} = \pi_{\text{Expert}}(\mathbf{v}_{\text{slow}}, \mathbf{O}_{\text{sen}}, \mathbf{O}_{\text{prop}}).
\end{equation}
In contrast, our objective is to learn a policy that enables the VLM to generate multi-timescale latent vectors, which directly guide the action expert to produce actions. We formulate this as:
\begin{equation}
    \mathcal{V}_{\text{f2s}} = \pi_{\text{VLM}}(\mathbf{O}_{\text{vis}}, \mathbf{L}_{\text{task}}, \mathbf{O}_{\text{sen}}, \mathbf{O}_{\text{prop}}),
\end{equation}
\begin{equation}
    \mathbf{A} = \pi_{\text{Expert}}(\mathcal{V}_{\text{f2s}}).
\end{equation}
Here, the multi-timescale latent vectors $\mathcal{V}_{\text{f2s}}$ spans slow-to-fast features, represented as a set $\{\mathbf{v}_k\}_{k=0}^{K-1}$, where $K$ denotes the number of distinct temporal scales integrated within the unified backbone.

\begin{figure*}[t]
    \centering
\includegraphics[width=1\textwidth]{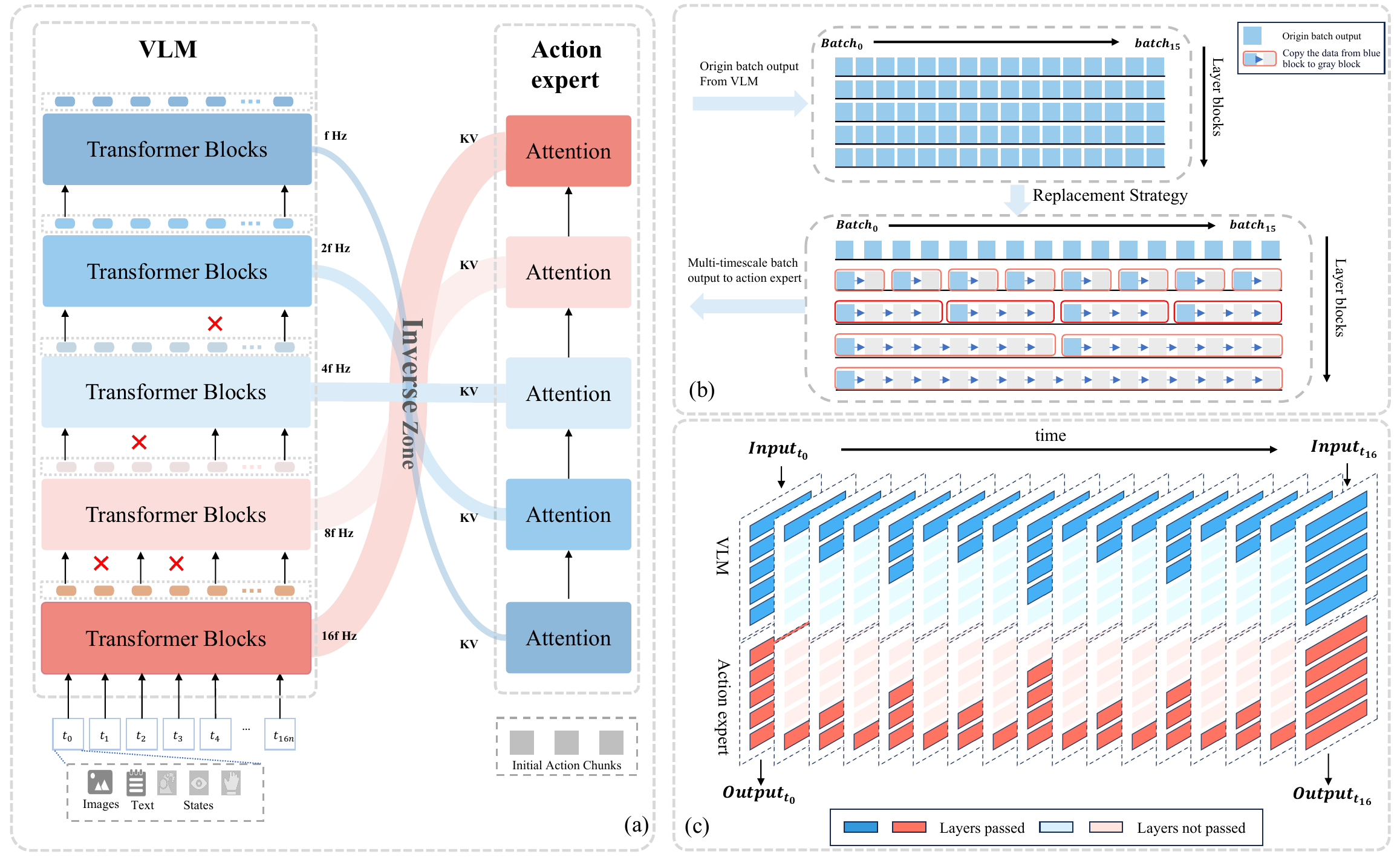}
    \caption{\textbf{Overview of our proposed \modelname.} In (a) We divide the VLM into multiple frequency groups, with each group operating at a fixed frequency. In (b), we illustrate how a batch performs temporal substitution under VLA parallel training. In (c), we intuitively illustrate the execution frequencies of different layers in one inference loop, showing a theoretical speedup of about $2.6\times$.}
    \label{fig:02model}
    \vspace{-25pt}
\end{figure*}

\subsection{Framework}
\label{subsec:architecture}
\noindent\textbf{Overall Architecture.}
To realize the policy formulation outlined in \Cref{subsec:problem}, we employ a hierarchical architecture aligned with mainstream VLA models~\cite{bjorck2025gr00t, li2026global, du2025himoe, li2025hamster, han2024dual, bu2024towards}. In this framework, the action expert leverages latent representations from each layer of the VLM, thereby guaranteeing it receives sufficiently rich information. Inspired by OpenVLA~\cite{kim2024openvla} and VLA-Adapter~\cite{wang2025vla}, we utilize DINOv2~\cite{oquab2023dinov2} and SigLIP~\cite{zhai2023sigmoid} as visual encoders to generate patch embeddings. These are subsequently concatenated with tokenized text instructions and processed by a pre-trained Large Language Model, such as LLaMA2~\cite{touvron2023llama} or Qwen~\cite{bai2023qwen}, to achieve efficient multimodal encoding.
Specifically, we extract hidden states from all intermediate layers of the LLM, serving as latent vectors with different frequencies that guide the action expert.

Our action expert comprises a stack of transformer blocks that process noisy action priors as input. As these queries propagate through the network, they fuse with the corresponding latent vectors via a cross-attention mechanism. Finally, a lightweight MLP head projects the refined query representations into executable action chunks, enabling precise robot manipulation.

\noindent\textbf{Fast-to-Slow Architecture Design.}
Standard VLM forward passes do not inherently support multi-timescale latent vectors. To address this, we propose a Fast-to-Slow Architecture (FSA), an asynchronous layer-wise execution mechanism illustrated in \Cref{fig:02model}. Specifically, we stratify the VLM layers into $K$ groups, each assigned a distinct update frequency. Formally, let the $k$-th layer group be updated once every $n_k$ timesteps, where
\[
n_1 < n_2 < \cdots < n_K
\]
\textit{i.e.}, $n_k$ increases with the network depth. 
Concretely, we set $n_1 = 1$, for the first layer group to ensure fastest-frequency perception, whereas the last group is assigned the maximum interval $n_K = N$, implying an update rate of once per 
$N$ timesteps.
Let $h_k^{(t)}$ denote the output features of the $k$-th layer group at timestep $t$, and let $F_k(\cdot)$ be the transformation implemented by this group. Then the temporal update rule is given by
\begin{equation}
h_k^{(t)} =
\begin{cases}
F_k\big(h_{k-1}^{(t)}\big), & \text{if } t \equiv 0 \pmod{n_k},\\[4pt]
h_k^{(t-1)}, & \text{otherwise}.
\end{cases}
\end{equation}
Formally, at timestep $t$ the $k$-th group is recomputed only if $t$ is divisible by $n_k$. Otherwise, its output is retained from the most recent valid update, avoiding redundant computation.
And we can obtain the multi-timescale latent vector \[
\mathcal{V}^{(t)}_{\text{f2s}}
= \big\{\, h_k^{(t)} \mid k \in \mathcal{K} \,\big\},
\] at timestep $t$, where $\mathcal{K}$ mean the set of all layer index.

\noindent\textbf{Latent Vector Inversion.} Empirical results in \Cref{fig:insight} reveal that hidden state representations in deeper layers exhibit significantly higher temporal frequencies. We attribute this phenomenon to the fast-frequency supervision of ground-truth actions, as deeper layers are inherently closer to the fast-frequency action output. However, this trend conflicts with the requirements of our fast-to-slow architecture which necessitates progressively slower update frequencies for deeper layers to cache previous hidden state representation. To achieve this, we introduce a latent vector inversion mechanism as shown in \Cref{fig:02model}. Our analysis indicates that such rapid variations stem from the iterative refinement process near the action generation. Consequently, we propose inverting the interaction sequence of latent vectors between the LLM and the action expert. Specifically, initial noisy action proposals interact with deeper features, while refined action outputs interact with shallower latent vectors. This re-routing shifts the high-frequency dynamics to the shallower layers, allowing the deeper layers to remain stable and cacheable without compromising action responsiveness.

\subsection{Training}
\noindent\textbf{Temporal Batch.} To enable stable training of our asynchronous layer-wise execution, we introduce a temporal batch sampling strategy that preserves temporal coherence within each batch, while keep random across batches.  The procedure is detailed in \Cref{alg:temporal_batch}. Given a trajectory of length $T$, we first randomly select a contiguous temporal window from the full trajectory. Within this window, we randomly sample $K$ timesteps and sort them by time, so that the resulting sequence preserves causal dependencies while still introducing stochasticity for robustness. For trajectories shorter than $K$, we apply causal padding by repeating the final observation–action pair to keep the batch dimensions consistent. The sampled temporal data are then assembled into a training batch for training asynchronous layer-wise execution.

\begin{algorithm}[t]
\caption{Temporal Batch Sampling}
\label{alg:temporal_batch}
\begin{algorithmic}[1]
\Require Trajectory $\tau = \{(\mathbf{o}_t, \mathbf{a}_t)\}_{t=1}^T$, sub-trajectory length $K$, effective window size $W_{\max}$
\Ensure Temporally-ordered subset $\tau_{\text{sub}}$ of size $K$

\If{$T < K$}
    \State $\mathcal{I} \gets [1, 2, \dots, T, \underbrace{T, \dots, T}_{K-T \text{ times}}$] \Comment{padding}
    \State \Return $\{(\mathbf{o}_{t}, \mathbf{a}_{t}) \mid t \in \mathcal{I}\}$
\EndIf

\State $W \gets \min(T, W_{\max})$ \Comment{Effective window size}
\State $t_{\text{start}} \gets \text{Uniform}(0, T - W)$ \Comment{Random window start}

\State $\mathcal{C} \gets \{t_{\text{start}}, t_{\text{start}}+1, \dots, t_{\text{start}}+W-1\}$ \Comment{Candidate indices}
\State $\mathcal{I}_{\text{rand}} \gets \text{RandomSample}(\mathcal{C}, K)$ \Comment{Draw $K$ indices without replacement}
\State $\mathcal{I} \gets \text{Sort}(\mathcal{I}_{\text{rand}})$ \Comment{Restore temporal order}

\State \Return $\{(\mathbf{o}_{t}, \mathbf{a}_{t}) \mid t \in \mathcal{I}\}$
\end{algorithmic}
\end{algorithm}

\noindent\textbf{Frequency Feature Replacement Strategy.}
A naive implementation of Asynchronous Layer-wise Execution would process timesteps sequentially within a batch: at timestep $t$, only the layers whose update frequency divides $t$ are recomputed, while all other layers simply reuse their cached features from previous timesteps.
However, this design disables parallel computation, severely under-utilizing GPUs and increasing training time by orders of magnitude. In addition, dynamically skipping layer computation breaks the static computation graph assumed by modern deep learning frameworks, introducing considerable engineering complexity, hurting reproducibility, and making integration with standard training pipelines difficult.

To overcome these challenges, we propose Frequency Feature Replacement (FFR), a training-time feature replacement strategy that decouples computation parallelism from update asynchrony. The key idea is to first compute all layer features for all timesteps within a batch using a standard, fully parallel forward pass, and then retrospectively align each feature to its prescribed update frequency via a differentiable indexing operation. As illustrated in \Cref{fig:02model}(b), where rows correspond to outputs of different layers and columns represent batch elements ordered in time, the original fully computed feature map (top panel) is transformed into a multi-timescale output (bottom panel). This transformation is performed by grouping consecutive timesteps into blocks, highlighted by the red boxes, within which features are shared. Concretely, for a layer with update frequency $f$, the feature at the start of each block is copied to the subsequent $f-1$ timesteps; for example, if $f = 4$, the feature at $t = 0$ is reused for $t = 1, 2, 3$. In this way, slower layers effectively update less frequently, while preserving the structure required for fully parallel computation.

Formally, let $\mathbf{H} \in \mathbb{R}^{T \times F \times N \times C}$ denote the hidden states across $T$ timesteps and $F$ layers. 
Given timescales $\mathbf{f} \in \mathbb{N}^F$ of all layer, we compute the aligned features $\mathbf{H}'$ as:
\begin{equation}
    \mathbf{H}'[t, f, :, :] = \mathbf{H}\big[\lfloor t / f \rfloor \times f,\; f,\; :, :\big], \quad \forall t \in [0, T), f \in [0, F).
\end{equation}

FFR enables efficient training of asynchronous layer-wise execution by maintaining full batch parallelism during the forward pass while achieving frequency-aware update patterns through a lightweight post-processing step, without sacrificing model expressiveness or gradient fidelity.

\begin{algorithm}[t]
\caption{Frequency Feature Replacement Strategy}
\label{alg:freq_replace}
\begin{algorithmic}[1]
\Require
    Hidden states $\mathbf{H} \in \mathbb{R}^{T \times F \times N \times C}$,
    timescales $\mathbf{f} \in \mathbb{N}^F$
\Ensure
    Replaced hidden states $\mathbf{H}' \in \mathbb{R}^{T \times F \times N \times C}$

\State Initialize $\mathbf{H}'$ with the same shape as $\mathbf{H}$

\For{$i = 0$ to $F-1$} \Comment{timescale index}
    \State $f \gets \mathbf{f}[i]$
    \For{$t = 0$ to $T-1$} \Comment{time index}
        \State $k \gets \big\lfloor \frac{t}{f} \big\rfloor \times f$ \Comment{latest time $\le t$ where this timescale is updated}
        \State $\mathbf{H}'[t, i, :, :] \gets \mathbf{H}[k, i, :, :]$
    \EndFor
\EndFor

\State \Return $\mathbf{H}'$
\end{algorithmic}
\end{algorithm}

\noindent\textbf{Multi-Level Supervision.} 
The model may learn shortcuts through fast-frequency components introduced by the inversion operation. To mitigate this, we adopt multi-level supervision, where the loss is applied not only at the final layer but also as auxiliary losses on each frequency group within the action expert.

For lower-frequency groups, whose hidden states are cached and updated less often, this means they receive identical features but different ground-truth actions across timesteps. This encourages them to ignore high-frequency noise and instead learn stable, coarse-grained plans over longer horizons, while higher-frequency groups focus on fine-grained, immediate motor corrections.

The total training objective is defined as the average of losses over all frequency levels
$$\mathcal{L}_{\text{total}} = \frac{1}{K} \sum_{k=0}^{K-1} \mathcal{L}_{\text{L1}}(f(\mathbf{h}_k), \mathbf{A}_{\text{gt}}),$$
where $\mathbf{h}_k$ denotes the hidden state after the $k$-th frequency group in the action expert, and $f$ is a shared lightweight mlp to decode the action. This uniform averaging enforces a coarse-to-fine learning hierarchy across temporal scales.

\subsection{Inference}
\label{sec:inference}

During inference, we fully exploit the computational benefits of our asynchronous design by caching latent vectors in both the VLM and the Action Expert as shown in \Cref{fig:02model} (c). Because the model has been trained to produce consistent predictions when features are reused across timesteps, the inference  remains exactly aligned with training distribution, avoiding any train–test discrepancy.

For the VLM backbone, layers assigned to lower frequency groups are computed only once every $f_k$ timesteps, with their hidden states cached and reused at intermediate steps. This reduces the per-timestep transformer computation proportionally to the update frequency. 

The action expert inference is further accelerated through latent vector inversion. Slowly varying latent vectors that encode relatively stable semantic context are injected into early layers, while rapidly varying latent vectors that represent fine-grained sensory details are injected into later layers. As a result, the early layers, driven by slowly updating semantic inputs, can operate at a reduced update frequency, whereas only the final layers, which integrate high-frequency latent vectors with proprioceptive feedback, must be executed at every timestep.

Consequently, both the VLM and the action expert achieve coordinated acceleration, yielding an overall theoretical speedup of about $2.6\times$ while maintaining action quality and temporal responsiveness.
\section{Experiments}
\subsection{Simulation Experiments on LIBERO}
\begin{figure*}[t]
    \centering
\includegraphics[width=\textwidth]{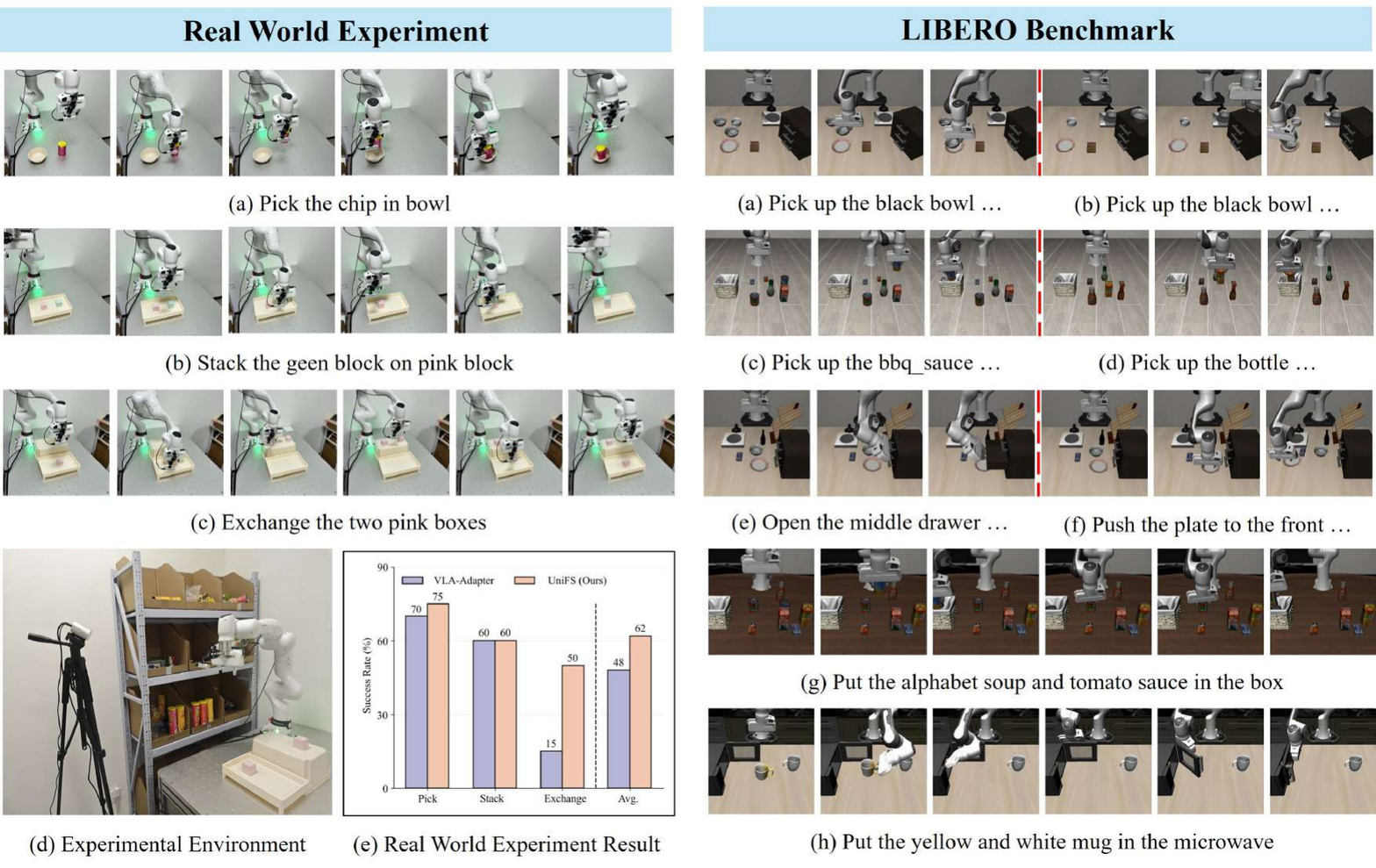}
    \caption{Visualization of tasks from the four LIBERO benchmark suites and three Real-World Franka scenarios.}
    \label{fig:exp}
    \vspace{-15pt}
\end{figure*}

\textbf{LIBERO Benchmark.}
As shown in \Cref{fig:exp} (right), we evaluate our model on the LIBERO~\cite{liu2023libero}, a comprehensive lifelong robotic learning benchmark with 130 tasks spanning four suites: (1) LIBERO-Spatial (a, b) evaluates spatial reasoning by varying camera viewpoints and object positions while keeping task semantics constant; (2) LIBERO-Object (c, d) tests object generalization by introducing diverse object instances with similar affordances; (3) LIBERO-Goal (e, f) assesses goal-conditioned learning by requiring different end states under similar initial configurations; and (4) LIBERO-10 (g, h), also called LIBERO-Long, is a challenging sequential multi-task suite requiring agents to master 10 distinct manipulation skills without catastrophic forgetting. 
This simulation environment is built on robosuite~\cite{zhu2020robosuite} with a Franka Panda robot model, designed to evaluate lifelong learning capabilities in manipulation tasks. 

\noindent\textbf{Implementation Details.} We adopt Qwen2.5-0.5B as the language backbone, while the action expert is randomly initialized. All experiments are conducted on 4 NVIDIA A100 GPUs using Distributed Data Parallel training. We use a per-GPU mini-batch size of 16, resulting in a global batch size of 64. We divide the VLM into 5 groups with time scales {1, 2, 4, 8, 16} and an effective window size of 128. The model is optimized with AdamW using a constant learning rate of $1\times10^{-5}$. For data loading, we utilize TensorFlow Datasets TFDS with a shuffle buffer size of $10000$. On LIBERO benchmark, the model takes as input both a third-person image and a wrist-mounted camera image together with the proprioceptive state of the robot, and outputs an action chunk of length 8 at each prediction step.

\noindent\textbf{Main Results.}
We report the success rate in \Cref{tab:libero}. Our \modelname\ achieves state-of-the-art results with an average success rate of $98.3\%$. For each task, \modelname\ obtains $99.6\%$, $99.6\%$, $98.1\%$, and $95.6\%$ on LIBERO-Spatial, LIBERO-Object, LIBERO-Goal, and LIBERO-Long, respectively. Compared with the baseline VLA-Adapter$^\dagger$ (modified to match our settings, \textit{e.g.}, temporal batch size), training with the proposed frequency feature replacement strategy yields a $2.5\%$ improvement and is also more efficient, demonstrating the effectiveness of our fast-to-slow architecture.

\begin{table*}[h]
\centering
\caption{\textbf{Performance on the LIBERO benchmark.} Success
rates (\%) are reported across four suites. \textbf{Bold} is the best performance}
\footnotesize
\setlength{\tabcolsep}{5pt}
\renewcommand{\arraystretch}{1.1}
\setlength{\tabcolsep}{10pt}
\resizebox{\columnwidth}{!}{
\begin{tabular}{l|l|cccc|c}
\toprule
\textbf{Architecture} & \textbf{Method} & \textbf{Spatial} & \textbf{Object} & \textbf{Goal} & \textbf{Long} & \textbf{Avg.} \\
\midrule

{Non-VLA}
  & Diffusion Policy$^\dagger$~\cite{chi2025diffusion} & -- & 78.3 & 92.5 & 68.3 & 72.4 \\
  & UniAct~\cite{zheng2025universal} 	        & 77.0 & 87.0 & 77.0 & 70.0 & 74.0\\
\midrule

\multirow{5}{*}{Standard VLA}
  & $\pi_0$~\cite{black2024pi0}               & 96.8 & 98.8 & 95.8 & 85.2 & 94.2 \\
  & $\pi_{0.5}$~\cite{black2024pi0}           & 98.0 & 97.8 & 95.6 & 85.8 & 94.3 \\
  & EO-1~\cite{qu2025eo}                      & 99.7 & 99.8 & 99.2 & 94.8 & 98.2 \\
  & X-VLA~\cite{zheng2025x}                   & 98.2 & 98.6 & 97.8 & 97.6 & 98.1 \\
  & OpenVLA~\cite{kim2024openvla}             & 84.7 & 88.4 & 79.2 & 53.7 & 76.5 \\
\midrule

\multirow{8}{*}{Enhanced VLA}
  & CoT-VLA~\cite{zhao2025cot}                & 87.5 & 91.6 & 87.6 & 69.0 & 81.1 \\
  & WorldVLA~\cite{cen2025worldvla}           & 87.6 & 96.2 & 83.4 & 60.0 & 81.8 \\
  & MemoryVLA~\cite{shi2025memoryvla}         & 98.4 & 98.4 & 96.4 & 93.4 & 96.5 \\
  & CogVLA~\cite{li2025cogvla}                & 98.5 & 98.8 & 96.5 & 95.2 & 97.4 \\
  & OpenVLA-OFT~\cite{kim2025fine}            & 96.2 & 98.3 & 96.2 & 94.5 & 95.3 \\
  & SmolVLA~\cite{shukor2025smolvla}          & 93.0 & 94.0 & 91.0 & 77.0 & 88.8 \\
  & GraspVLA~\cite{deng2025graspvla}          & --   & 94.1 & 91.2 & 82.0 & 89.1 \\
  & SpatialVLA~\cite{qu2025spatialvla}        & 88.2 & 89.9 & 78.6 & 55.5 & 78.1 \\
\midrule

\multirow{5}{*}{Fast-Slow VLA}
  & $\pi_0$-FAST~\cite{pertsch2025fast}       & 96.4 & 96.8 & 88.6 & 60.2 & 85.5 \\
  & VLA-Cache~\cite{xu2025vla}                & 98.3 & 97.5 & 98.3 & 95.4 & 97.4 \\
  & PD-VLA~\cite{song2025pd}                  & 95.5 & 96.7 & 94.9 & 91.7 & 94.7 \\
  & GR00T N1~\cite{bjorck2025gr00t}           & 94.4 & 97.6 & 93.0 & 90.6 & 93.9 \\
  & SP-VLA~\cite{li2025sp}                    & 84.4 & 85.6 & 75.4 & 74.9 & 80.1 \\
\midrule

\rowcolor[gray]{0.9}
  & VLA-Adapter~\cite{wang2025vla}$^\dagger$  & 97.2 & 98.8 & 93.4 & 93.6 & 95.8 \\
\rowcolor[gray]{0.9}
\multirow{-2}{*}{\textbf{Ours}}
  & \textbf{\modelname}                       & \textbf{99.6} & \textbf{99.6} & \textbf{98.2} & \textbf{95.6} & \textbf{98.3} \\
\bottomrule
\end{tabular}
}
\label{tab:libero}
\vspace{-10pt}
\end{table*}

\noindent\textbf{Efficiency Experiment.} 
\modelname\ runs faster than baselines as its asynchronous layer-wise execution. \Cref{tab:efficiency} compares with OpenVLA, OpenVLA-OFT, and VLA-Adapter. Our \modelname\ achieves an average latency of only $17.8$ ms, which is much faster than VLA-Adapter. 

\begin{table}[t]
\centering
\footnotesize
\caption{
Efficiency comparison of OpenVLA, OpenVLA-OFT, VLA-Adapter, and \modelname\ running with lowest latency of $12.3$ ms and largest latency of $32.6$ ms. In a whole period, \modelname\ achieves an average latency of $17.8$ ms.}
\label{tab:efficiency}
\resizebox{\columnwidth}{!}{
\setlength{\tabcolsep}{10pt}
\begin{tabular}{lccc|ccc}
\toprule
\multirow{2}{*}{\textbf{Metric}} & 
\multirow{2}{*}{OpenVLA} & 
\multirow{2}{*}{OpenVLA-OFT} & 
\multirow{2}{*}{VLA-Adapter} & 
\multicolumn{3}{c}{\textbf{UniFS (Ours)}} \\
\cmidrule(lr){5-7}
 & & & & Fastest & Mean & Slowest \\
\midrule
Throughput (Hz) $\uparrow$ & 4.2   & 109.7 & 219.2 & 650.4 & \textbf{449.4} & 245.4 \\
Latency (ms) $\downarrow$  & 239.6 & 72.9  & 36.5  & 12.3  & \textbf{17.8}  & 32.6  \\
\bottomrule
\end{tabular}
}
\end{table}

\subsection{Ablation Studies}
\textbf{Ablation study on effectiveness.}
We firstly investigate the contribution of each proposed component within \modelname. Starting from the base model, we incrementally introduce three key components: (1) Fast-to-Slow Architecture (\textit{e.g.}, FSA); (2) Latent Vector Inversion (\textit{e.g.}, LVI); and (3) Multi-Level Supervision (\textit{e.g.}, MLS). The success rates are reported in Table~\ref{tab:ablation}. With all components enabled, \modelname\ achieves an overall improvement of $2.5\%$, demonstrating the effectiveness of our proposed fast-to-slow architecture.

\begin{table}[t]
\centering
\footnotesize
\caption{Ablation study of \modelname\ components, including (1) Fast-to-Slow Layer; (2) Latent Vector Inversion and (3) Multi-Level Supervision. With all three components, we achieve the best results of $98.3\%$ success rate with $17.8$ ms latency.}
\label{tab:ablation}
\resizebox{\columnwidth}{!}{
\setlength{\tabcolsep}{10pt}
\renewcommand{\arraystretch}{1.15}
\begin{tabular}{ccc|cccc|c|c}
\toprule
\multicolumn{3}{c|}{\textbf{Components}} 
  & \multicolumn{5}{c|}{\textbf{Success Rate (\%)}} 
  & \multicolumn{1}{c}{\textbf{Latency (ms)}} \\
\cmidrule(lr){1-3} \cmidrule(lr){4-8}  \cmidrule(lr){9-9}
\textbf{FSA} & \textbf{LVI} & \textbf{MLS}
  & Spatial & Object & Goal & Long 
  & Avg.
  & Avg.\\
\midrule
$\times$ & $\times$ & $\times$
  & 97.2 & 98.8 & 93.4 & 93.6 
  & 95.8
  & 36.5 \\
\midrule
$\checkmark$ & $\times$ & $\times$
  & 83.4 & 93.2 & 76.0 & 28.4 
  & 70.3
  & 22.6 \\
$\checkmark$ & $\checkmark$ & $\times$
  & 98.8 & 98.8 & 88.4 & 91.0 
  & 94.3
  & 17.8 \\
\rowcolor[gray]{0.92}
$\checkmark$ & $\checkmark$ & $\checkmark$
  & \textbf{99.6} & \textbf{99.6} & \textbf{98.2} & \textbf{95.6} 
  & \textbf{98.3}
  & \textbf{17.8} \\
\bottomrule
\end{tabular}
}
\vspace{-15pt}
\end{table}

\textbf{Baseline.} The baseline builds upon VLA-Adapter~\cite{wang2025vla} which already achieves an average of $95.8\%$, providing a reliable foundation for subsequent analysis.

\textbf{+FSA.} When only enabling the FSA, causes a performance drop to $70.3\%$, especially drop from $93.6\%$ to $28.0\%$ on LIBERO-10. We attribute this instability to a frequency mismatch between the multi-scale latent vectors and the action expert, which directly limits the effectiveness of modeling the fast-to-slow layer.

\textbf{+LVI.} When we invert the multi-scale latent vectors to make them align with the action expert, the average success rate increases to $94.3\%$, which is comparable to baseline. This shows the crucial role of LVI in fast-to-slow layer.

\textbf{+MLS.} As mentioned above, inverting the latent vectors may create shortcuts that prevent the shadow layer from being properly trained. Introducing multi-layer supervision, \modelname\ achieves better results across all tasks.

\begin{wrapfigure}{r}{0.5\textwidth}
    \vspace{-15pt}
    \centering
    \includegraphics[width=0.5\textwidth]{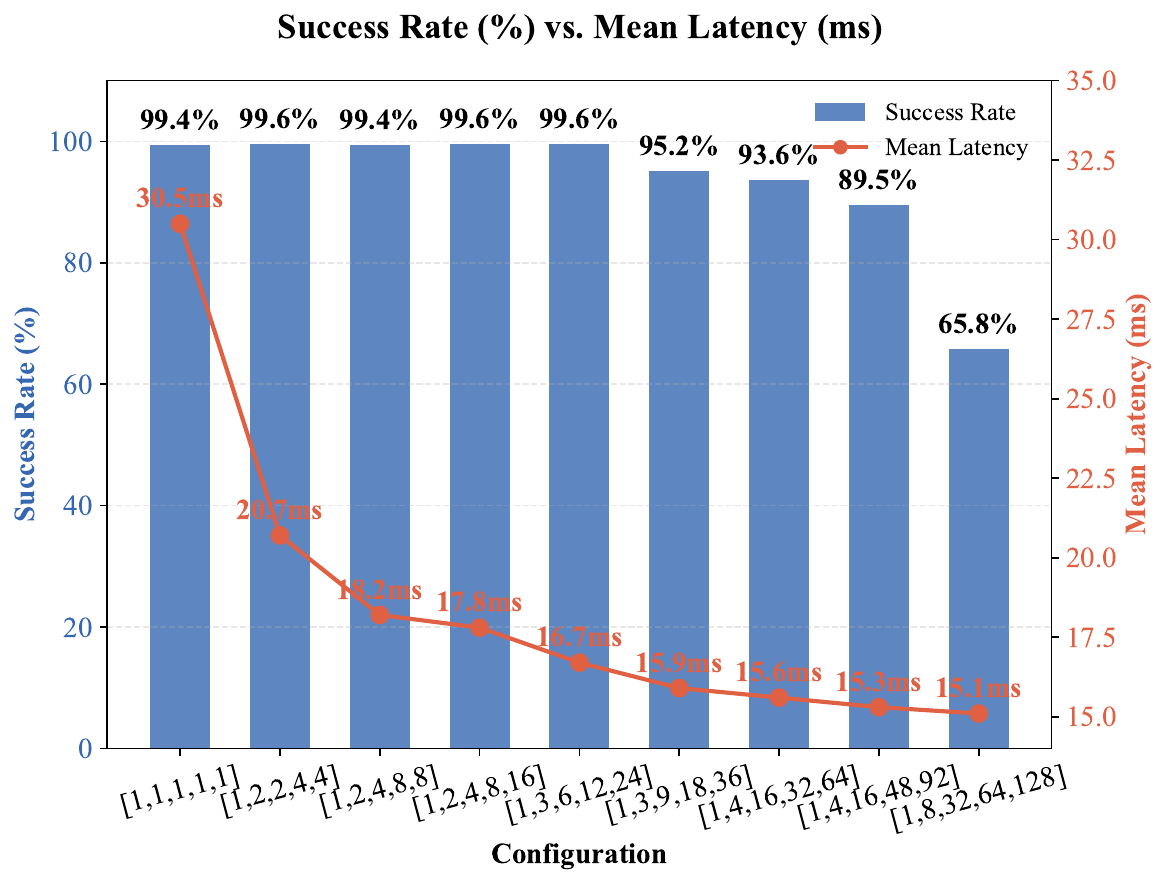}
    \caption{\textbf{Ablation study} on sampling frequency configurations.}
    \label{fig:ablation_freq}
    \vspace{-15pt}
\end{wrapfigure}
\noindent\textbf{Ablation study on efficiency.} 
Benefiting from the temporal batch sampling strategy with random temporal sampling, the frequency for each group can also be customized during inference, although it is fixed to half that of the preceding group during training. We characterize each configuration by the maximum-to-minimum sampling frequency ratio across groups, ranging from $1\times$ (\textit{e.g.}, $[1,1,1,1,1]$) to $128\times$ (\textit{e.g.}, $[1,8,32,64,128]$). The results on LIBERO-Spatial are shown in \Cref{fig:ablation_freq}. As illustrated, when the maximum-to-minimum sampling frequency ratio does not exceed $24\times$, there is negligible impact on performance. Beyond this threshold, larger ratios lead to performance degradation, yet offer only marginal latency gains. This demonstrates that \modelname\ is robust to varying frequency configurations that differ from the training setup; notably, even uniform sampling (\textit{i.e.}, no frequency) works without performance degradation.

\subsection{Deployment on Real Robot}
\noindent\textbf{Environment Setup.}
To validate the real-world applicability of \modelname, we deploy it on a Franka Research 3 (FR3) platform, as shown in Figure~\ref{fig:exp}(d). The system comprises a 7-DOF collaborative arm equipped with two complementary perception modules: an Orbbec RGB-D camera for scene-level understanding and an Intel RealSense camera mounted at the end-effector for precise manipulation perception. We evaluate \modelname\ on three tasks of increasing complexity, as illustrated in Figure~\ref{fig:exp}(a)--(c): (a) Pick, a fundamental object grasping task; (b) Stack, requiring precise spatial reasoning for object placement; and (c) Exchange Boxes, a short-horizon memory-aware task that requires the robot to reason over prior actions and object states.

\noindent\textbf{Pick and Stack.} For these two foundational manipulation tasks, \modelname\ achieves success rates of $75\%$ and $60\%$, respectively in pick and stack, which are comparable to our baseline model without frequency. 

\noindent\textbf{Exchange Boxes.} This task presents a significantly greater challenge, requiring the robot to exchange two cubes between separate containers. It necessitates short-term memory capabilities, as the correct action is ambiguous in many states when relying solely on the current observation. Despite this complexity, \modelname\ achieves a $50\%$ success rate.
\section{Conclusion}
\label{sec:conclusion}
In this work, we introduced \modelname, a novel unified fast-to-slow architecture biologically inspired  by multi-timescale processing mechanism of human brain to overcome the frequency dilemma and integration challenges in mainstream Fast-Slow Dual System VLA models. Through Fast-to-Slow Architecture, Latent Vector Inversion, and Multi-Level Supervision, \modelname\ resolves these limitations through a unified architecture that produces multi-frequency latent vectors from the VLM, facilitates effective information transfer to the action expert, and inherently provides the system with implicit short-term memory. Extensive experiments on the LIBERO benchmark demonstrate that \modelname\ achieves state-of-the-art performance and speed with an average success rate of 98.3\%, with only $17.8$ ms latency.

\noindent\textbf{Limitations and Future Work.}
Despite the substantial inference acceleration, \modelname\ still has several limitations. The training process does not benefit from computational speedup, as the Frequency Feature Replacement strategy requires full parallel forward passes to maintain GPU utilization. Stabilizing multi-frequency learning further demands relatively large batch sizes to ensure sufficient temporal diversity within each update. In addition, the fast-to-slow architecture and latent vector inversion introduce distribution shifts that may interfere with the semantic knowledge encoded in pre-trained backbones, necessitating careful fine-tuning to avoid performance degradation. In future, we will first extend \modelname\ to more advanced backbone architectures (\textit{e.g.}, $\pi_\text{0}$), and then systematically study how such adaptations affect the underlying pre-trained representations. Building on these insights, we will further integrate \modelname\ into state-of-the-art foundation models and conduct large-scale pre-training to better preserve, and potentially enhance, their semantic capacity.

\bibliographystyle{splncs04}
\bibliography{main}


\clearpage
\setcounter{page}{1}
\appendix

\section{Detailed Architecture} 
\label{sec:architecture}

As shown in \Cref{tab:latency_breakdown}, in our implementation, we find that the vision encoder is more time-consuming than LLM , so both vision encoder and LLM backbone are structured as fast-to-slow architecture. \Cref{tab:model_config} presents the comprehensive configuration settings for our \modelname, which is built on VLA-Adapter. 

The visual inputs are processed by both DINOv2 and SigLIP encoders, where the hidden states are extracted at relative frequencies of 16$f$, 8$f$ and 4$f$, where $f$ means the base frequency in \modelname. The outputs of vision encoders are processed by Qwen2.5 along with language instruction, where the hidden states are extracted at relative frequencies of 4$f$, 2$f$ and $f$. Notably, the frequency of 4$f$ applies to both vision encoder and LLM backbone. 
\begin{table}[h]
\centering
\caption{Inference time of each module in \modelname.}
\label{tab:latency_breakdown}
\resizebox{0.4\columnwidth}{!}{%
\begin{tabular}{lccc}
\toprule
\textbf{Module} & \textbf{Time (ms)} & \textbf{Ratio (\%)} \\
\midrule
Vision Encoder     & 13.5             & 41.4 \\
LLM                & 9.6              & 29.4 \\
Action Expert      & 9.5              & 29.2 \\
\midrule
\textbf{Total}     & \textbf{32.6}    & \textbf{100.0} \\
\bottomrule
\end{tabular}%
}
\vspace{-10pt}
\end{table}

The action expert of \modelname\ consists of 24 transformer layers and selects a total of 24 hidden states from specific layers across the vision encoder and LLM backbone to serve as fast-to-slow latent vectors to guide action generation. 

\Cref{tab:model_config_freq} outlines the layer-wise frequency assignments, while \Cref{tab:model_config_layer} specifies the feature selection strategy for the action expert. Specifically, we extract output features from the terminal layers within each frequency group. Additionally, we incorporate features from intermediate LLM layers, as empirical evidence suggests they contribute significantly to representation quality and overall performance.
This selection ensures that the action expert receives a rich mixture of fast-to-slow features.

\begin{table}[t]
\centering
\caption{\textbf{Model Configuration and Latent Vector Selection Strategy.} 
(a) Relative frequency configuration of layers for both vision encoders and LLM backbone. A higher relative frequency indicates more frequent hidden state updating. 
(b) Hidden states of specific layers with different frequency selected to the action expert module.}
\label{tab:model_config}

\begin{subtable}[c]{0.48\textwidth}
\centering
\caption{Relative frequency per layer of both vision encoder and LLM backbone.}
\label{tab:model_config_freq}
\setlength{\tabcolsep}{4pt}
\renewcommand{\arraystretch}{1.1} 
\resizebox{\linewidth}{!}{%
\begin{tabular}{@{} l | l l c @{}} 
\toprule
\textbf{Module} & \textbf{Model} & \textbf{Layer Range} & \textbf{Frequency} \\
\midrule
\multirow{6}{*}{Vision} 
    & \multirow{3}{*}{DINO v2} & Layers 0--11  & 16$f$  \\
    &                          & Layers 12--17 & 8$f$  \\
    &                          & Layers 18--23 & 4$f$  \\
\cmidrule(l){2-4} 
    & \multirow{3}{*}{SigLIP}  & Layers 0--11  & 16$f$  \\
    &                          & Layers 12--17 & 8$f$  \\
    &                          & Layers 18--27 & 4$f$  \\
\midrule
\multirow{3}{*}{LLM} 
    & \multirow{3}{*}{Qwen2.5} & Layers 0--2   & 4$f$  \\
    &                          & Layers 3--11  & 2$f$  \\
    &                          & Layers 12--23 & $f$ \\
\bottomrule
\end{tabular}%
}
\end{subtable}
\hfill
\begin{subtable}[c]{0.48\textwidth}
\centering
\caption{hidden states of selected layers with specific frequency for action expert}
\label{tab:model_config_layer}
\setlength{\tabcolsep}{4pt}
\renewcommand{\arraystretch}{1.1}
\resizebox{\linewidth}{!}{%
\begin{tabular}{@{} l | l c @{}}
\toprule
\textbf{Module} & \textbf{Selected Layers} & \textbf{ Frequency} \\
\midrule
\multirow{2}{*}{Vision} & Layers 7--11    & 16$f$  \\
                                & Layers 13--17   & 8$f$  \\
\midrule
\multirow{3}{*}{LLM}    & Layers 0--2     & 4$f$  \\
                                & Layers 8--11    & 2$f$  \\
                                & Layers 12--15, 21--23 & $f$ \\
\bottomrule
\end{tabular}%
}
\end{subtable}
\vspace{-15pt}
\end{table}

\section{More Analysis}
\textbf{Layer-wise temporal dynamics.} 
To quantify the temporal variability of hidden states, we compute the cosine distance between consecutive time steps for each LLM layer of our proposed \modelname. Specifically, for a hidden state $h_t$ at time $t$, the cosine distance is defined as $1 - \frac{h_t \cdot h_{t+1}}{\|h_t\| \|h_{t+1}\|}$, measuring angular differences in the latent space where smaller values indicate more stable representations.

\begin{figure*}[h]
    \centering
    \includegraphics[width=\textwidth]{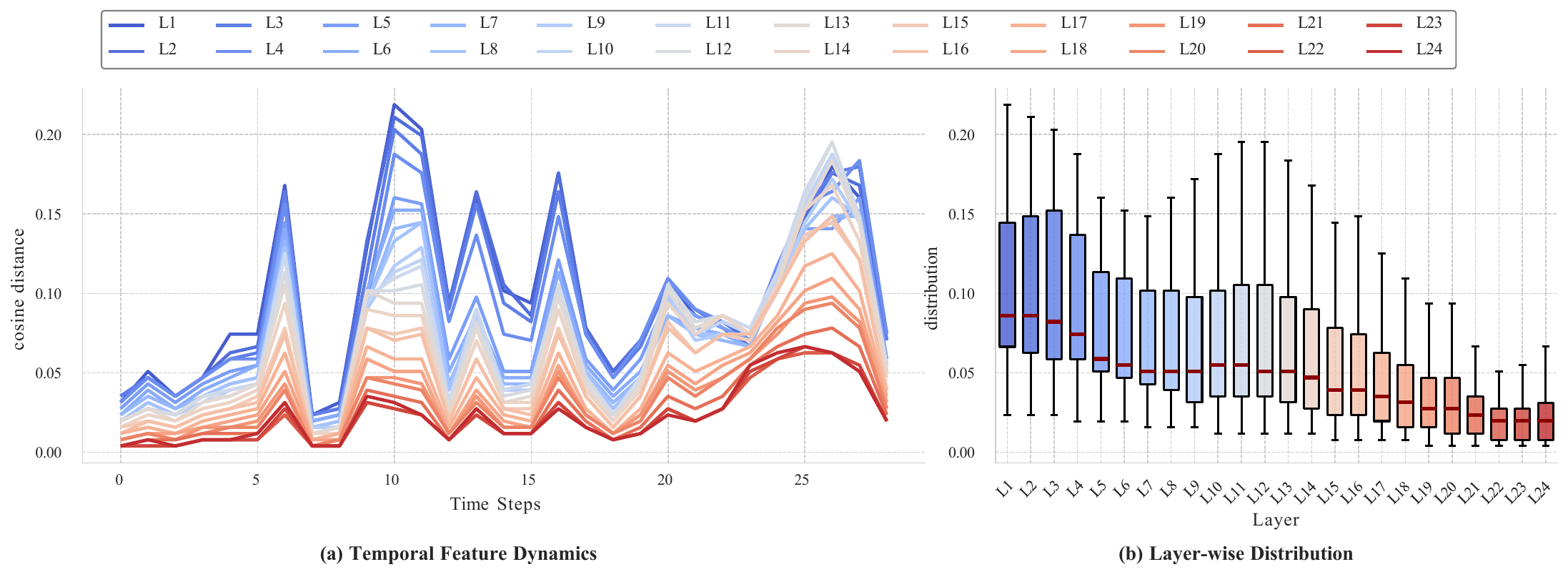}
    \caption{\textbf{Layer-wise temporal feature dynamics of \modelname.} 
    (a) Cosine distance between consecutive time steps per layer, with colors indicating layer depth (blue: shallow, red: deep). shallower layers fluctuate more while deeper layers remain stable. (b) Per layer distribution of cosine distances across time steps. Decreasing variance and     medians from shallower to deeper layers align with our fast-to-slow architecture.}
    \label{fig:temporal_uniFS}
    \vspace{-20pt}
\end{figure*}

\Cref{fig:insight}(a) (b) show the model without fast-to-slow architecture, where deeper layers demonstrate greater temporal variability with frequent fluctuations, while shallower layers maintain greater stability. We also show this variability of \modelname\ with the fast-to-slow architecture in \Cref{fig:temporal_uniFS}(a) and quantify in \Cref{fig:temporal_uniFS}(b), showing the distribution of cosine distances for each layer across all time steps. Each box plot summarizes temporal variation via median, interquartile range, and spread. Within \modelname\ shows an opposite observation. the shallower layers exhibit wider distributions with higher medians, indicating sensitivity to visual changes, while deeper layers show compact distributions, indicating stable semantic features.

We argue that this phenomenon is due to the latent vector inversion described in \Cref{subsec:architecture}. Specifically, shallower layers are closer to high-frequency, fine-grained actions, thereby requiring high-frequency features to adjust. 
This requirement directly supports our fast-to-slow architecture, which is designed to enable faster variations in shallow layers to match these high-frequency dynamics. With this design, we can reuse features and update them only at their corresponding frequencies, thereby accelerating inference.

\begin{figure*}[h]
    \centering
\includegraphics[width=0.8\textwidth]{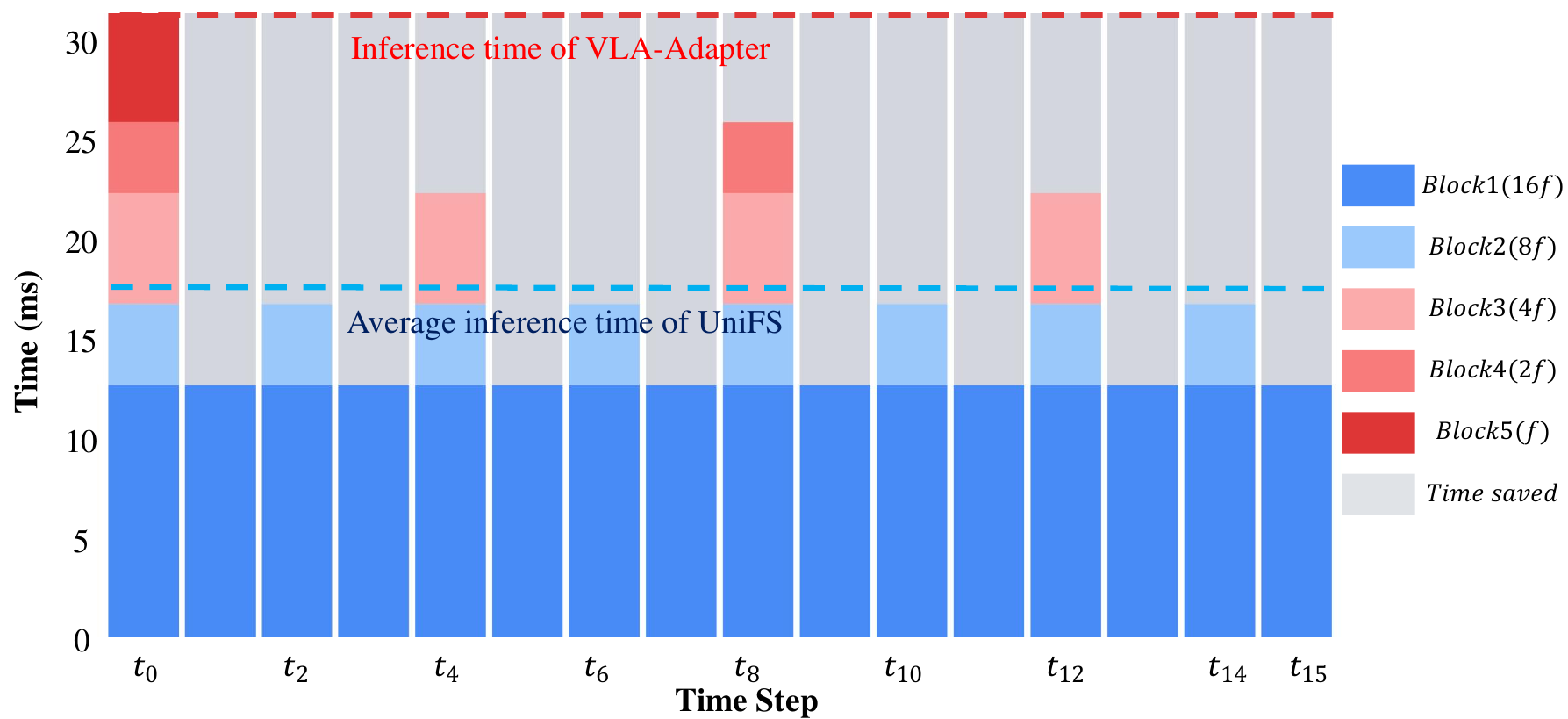}
    \caption{\textbf{Timing analysis.} We visualize the inference time of UniFS across continuous timesteps in Libero-Spatial suite.}
    \label{fig:timing}
    \vspace{-15pt}
\end{figure*}

\noindent\textbf{Timing analysis.} 
In \Cref{fig:timing}, we present the per-step inference time  for Libero-Spatial, revealing how the fast-to-slow scheduling strategy of \modelname\ achieves computational efficiency. \textit{Block1} (16$f$) executes at every step with minimal inference time (13.3~ms), where $f$ denotes the base inference frequency, establishing a consistent shallow baseline. To improve prediction accuracy within the time budget, \textit{Block2}--\textit{Block5} (8$f$, 4$f$, 2$f$, $f$) are selectively activated, with each layer addition contributing approximately 4--5~ms. This selective activation pattern is evident in the figure, which $t_0$ activates all layers to perform full inference (reaching 32~ms), while others activate some part blocks, maintaining fast inference. Across a complete inference cycle ($t_0$--$t_{15}$), the average inference time is 17.8~ms, substantially lower than the full layer inference time of 32~ms. 

\section{Visualization of Results}

\begin{figure*}[h]
    \centering
\includegraphics[width=0.9\textwidth]{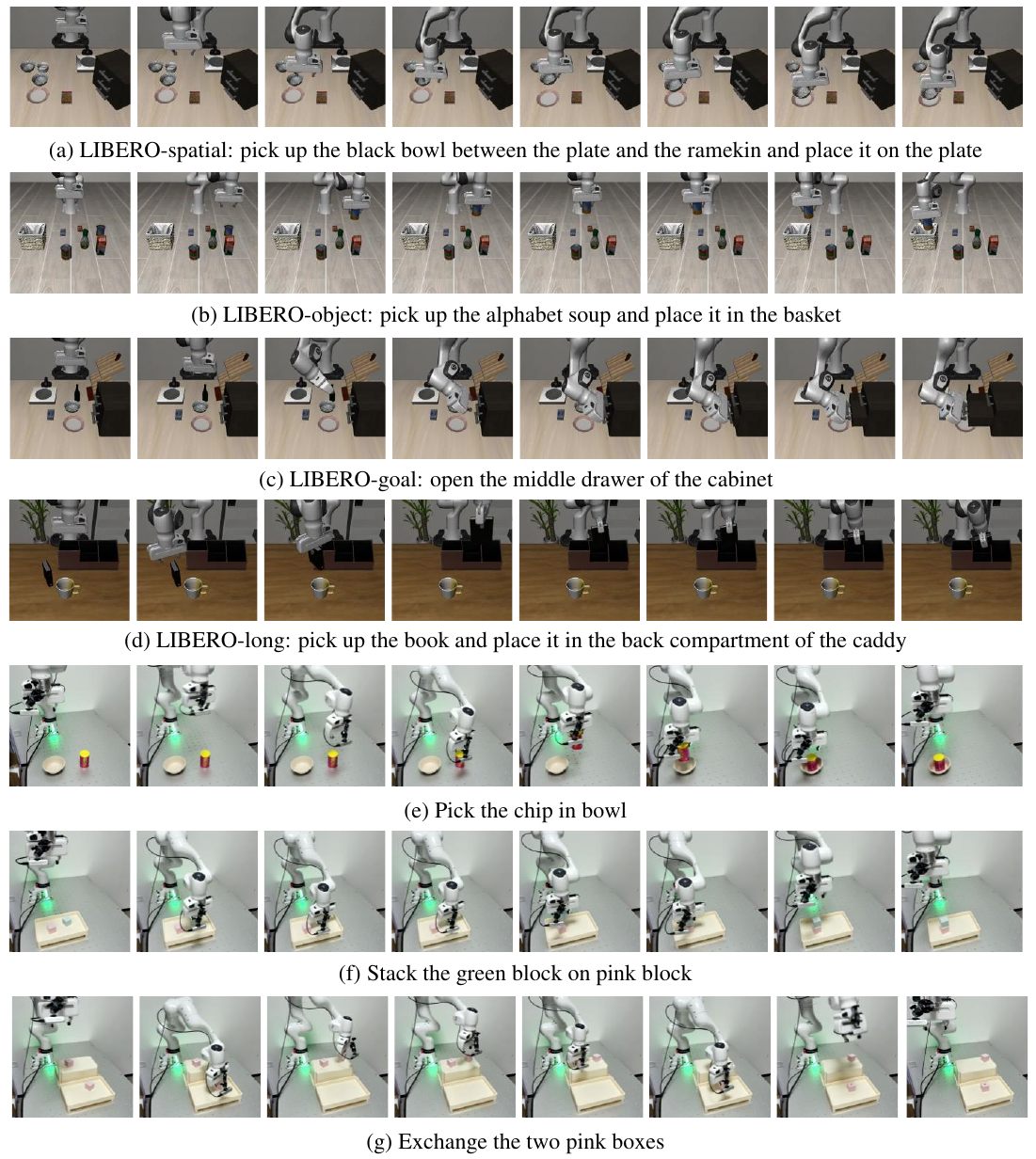}
    \caption{Qualitative visualizations of our proposed \modelname\ on both LIBERO and real-world experiments.}

    \label{fig:libero_franka}
    \vspace{-15pt}
\end{figure*}

\Cref{fig:libero_franka} presents qualitative visualizations of our \modelname\ on both the LIBERO benchmark and real-world Franka robot experiments. These visualizations demonstrate that our approach consistently achieves precise and robust manipulation performance.
Specifically, in the task of \Cref{fig:libero_franka}(g), there is ambiguity in many states during inference. For instance, after the robot picks the cube from the lower container to the higher container, the robot would be confused about exchanging which cube to the lower container based solely on current observations without historical context. 
However, compared with VLA-Adapter without reusing features, \modelname\ achieves a higher success rate on this task by reusing historical, low-frequency hidden states. This superior performance demonstrates that \modelname\ equips more robust implicit memory capabilities.

\end{document}